\documentclass[10pt,twocolumn,letterpaper]{article}

\usepackage{iccv}
\usepackage{times}
\usepackage{epsfig}
\usepackage{graphicx}
\usepackage{amsmath}
\usepackage{amssymb}
\usepackage{comment}
\usepackage{subcaption}
\usepackage{stfloats}
\usepackage[utf8x]{inputenc} 

\usepackage[pagebackref=true,breaklinks=true,letterpaper=true,colorlinks,bookmarks=false]{hyperref}

\iccvfinalcopy 

\newcommand{\CORRSFULL}[1]{{symmetry-aware object correspondences\xspace}}
\newcommand{\CORR}[1]{{SOC}}
\newcommand{\CORRS}[1]{{SOCs}}

\ificcvfinal\fi
\begin{document}

\title{End-to-End CAD Model Retrieval and 9DoF Alignment in 3D Scans}

\author{
Armen Avetisyan \qquad Angela Dai \qquad \qquad Matthias Nie{\ss}ner
\vspace{0.2cm} \\ 
Technical University of Munich
}

\twocolumn[{%
\renewcommand\twocolumn[1][]{#1}%
\vspace{-0.6cm}
\maketitle

\begin{center}
\vspace{-0.6cm}
\includegraphics[width=\linewidth]{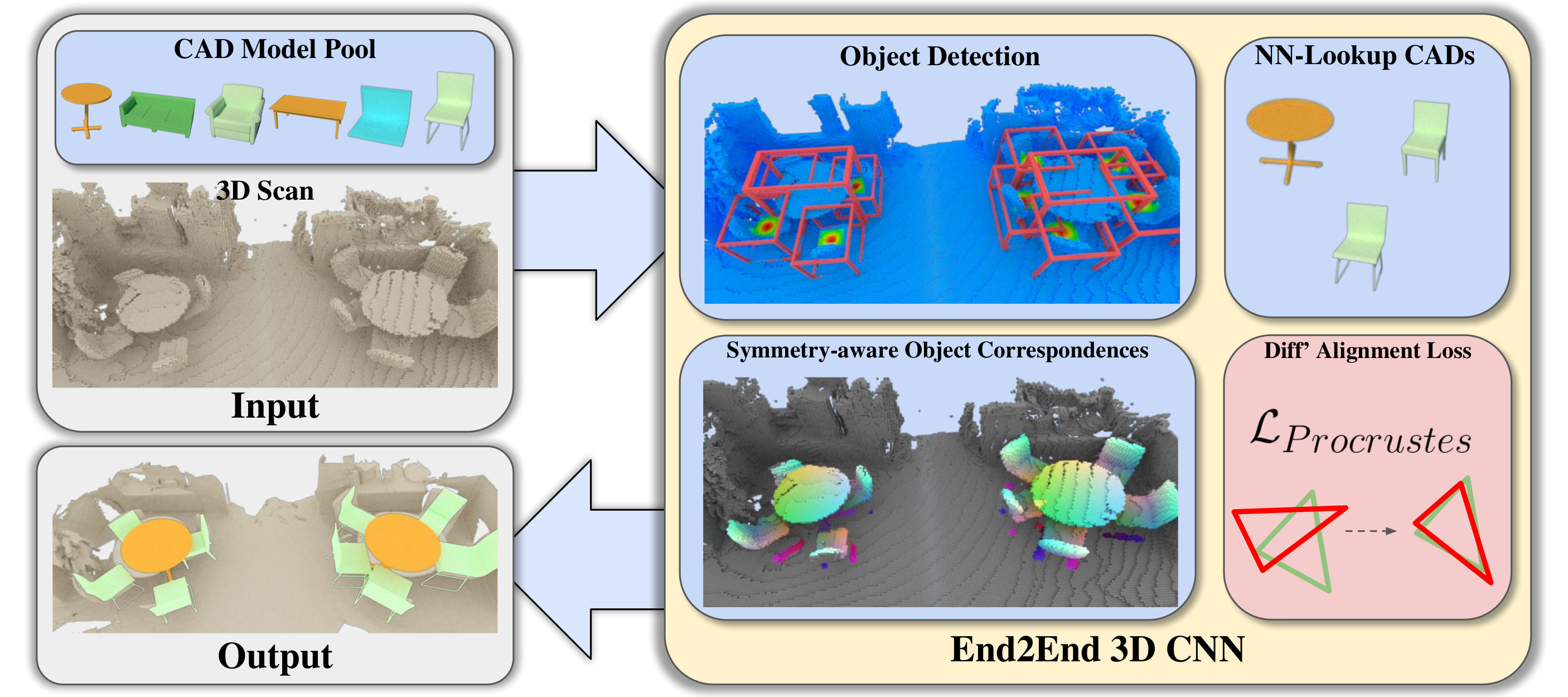}
\vspace{-0.6cm}
\captionof{figure}{
From a 3D scan and a set of CAD models, our method learns to predict 9DoF CAD model alignments to the objects of the scan, in a fully-convolutional, end-to-end fashion.
Our proposed 3D CNN first detects objects in the scan, then uses the regressed object bounding boxes to establish \CORRSFULL{} between a scan object and CAD model, which inform our differentiable Procrustes alignment loss, enabling learning of alignment-informed correspondences and producing CAD model alignment to a scan in a single forward pass.
}

\label{fig:teaser}
\end{center}
}]

\begin{abstract}
\vspace{-0.3cm}
We present a novel, end-to-end approach to align CAD models to an 3D scan of a scene, enabling transformation of a noisy, incomplete 3D scan to a compact, CAD reconstruction with clean, complete object geometry.
Our main contribution lies in formulating a differentiable Procrustes alignment that is paired with a symmetry-aware dense object correspondence prediction.
To simultaneously align CAD models to all the objects of a scanned scene, our approach detects object locations, then predicts symmetry-aware dense object correspondences between scan and CAD geometry in a unified object space, as well as a nearest neighbor CAD model, both of which are then used to inform a differentiable Procrustes alignment.
Our approach operates in a fully-convolutional fashion, enabling alignment of CAD models to the objects of a scan in a single forward pass. 
This enables our method to outperform state-of-the-art approaches by $19.04\%$ for CAD model alignment to scans, with  $\approx 250\times$ faster runtime than previous data-driven approaches.
\end{abstract}
\section{Introduction}
In recent years, RGB-D scanning and reconstruction has seen significant advances, driven by the increasing availability of commodity range sensors such as the Microsoft Kinect, Intel RealSense, or Google Tango.
State-of-the-art 3D reconstruction approaches can now achieve impressive capture and reconstruction of real-world environments~\cite{izadi2011kinectfusion,newcombe2011kinectfusion,niessner2013hashing,whelan2015elasticfusion,whelan2015elasticfusion,choi2015robust,dai2017bundlefusion}, spurring forth many potential applications of this digitization, such as content creation, or augmented or virtual reality.

Such advances in 3D scan reconstruction have nonetheless remained limited towards these use scenarios, due to geometric incompleteness, noise and oversmoothing, and lack of fine-scale sharp detail.
In particular, there is a notable contrast in such reconstructed scan geometry in comparison to the clean, sharp 3D models created by artists for visual and graphics applications.

With the increasing availability of synthetic CAD models~\cite{shapenet2015}, we have the opportunity to reconstruct a 3D scan through CAD model shape primitives; that is, finding and aligning similar CAD models from a database to each object in a scan.
Such a scan-to-CAD transformation enables construction of a clean, compact representation of a scene, more akin to artist-created 3D models to be consumed by mixed reality or design applications.
Here, a key challenge lies in finding and aligning similar CAD models to scanned objects, due to strong low-level differences between CAD model geometry (clean, complete) and scan geometry (noisy, incomplete).
Current approaches towards this problem thus often operate in a sparse correspondence-based fashion~\cite{li2015database,avetisyan2019scan2cad} in order to establish reasonable robustness under such differences.

Unfortunately, such approaches, in order to find and align CAD models to an input scan, thus involve several independent steps of correspondence finding, correspondence matching, and finally an optimization over potential matching correspondences for each candidate CAD model.
With such decoupled steps, there is a lack of feedback through the pipeline; e.g., correspondences can be learned, but they are not informed by the final alignment task.
In contrast, we propose to predict symmetry-aware dense object correspondences between scan and CADs in a global fashion.
For an input scan, we leverage a fully-convolutional 3D neural network to first detect object locations, and then from each object location predict a uniform set of dense object correspondences and object symmetry are predicted, along with a nearest neighbor CAD model; from these, we introduce a differentiable Procrustes alignment, producing a final set of CAD models and 9DoF alignments to the scan in an end-to-end fashion. 
Our approach outperforms state-of-the-art methods for CAD model alignment by $19.04\%$ for real-world 3D scans.

Our approach is the first, to the best of our knowledge, to present an end-to-end scan-to-CAD alignment, constructing a CAD model reconstruction of a scene in a single forward pass.
In summary, we propose an end-to-end approach for scan-to-CAD alignment featuring:
\begin{itemize}
    \item a novel differentiable Procrustes alignment loss, enabling end-to-end CAD model alignment to a 3D scan,
    \item symmetry-aware dense object correspondence prediction, enabling robust alignment even under various object symmetries, and
    \item CAD model alignment for a scan of a scene in a single forward pass, enabling very efficient runtime ($< 3$s on real-world scan evaluation)
\end{itemize}

\section{Related work}
\paragraph{RGB-D Scanning and Reconstruction}
3D scanning methods have a long research history across several communities, ranging from offline to real-time techniques. 
In particular, RGB-D scanning has become increasingly popular, due to the increasing availability of commodity range sensors.
A very popular reconstruction technique is the volumetric fusion approach by Curless and Levoy~\cite{curless1996volumetric}, which has been materialized in many real-time reconstruction frameworks such as KinectFusion~\cite{izadi2011kinectfusion,newcombe2011kinectfusion}, Voxel Hashing~\cite{niessner2013hashing} or BundleFusion~\cite{dai2017bundlefusion}, as well as in the context of state-of-the-art offline reconstruction methods~\cite{choi2015robust}.
An alternative to these voxel-based scene representations is based on surfels \cite{keller2013real}, that has been used by ElasticFusion \cite{whelan2015elasticfusion} to realize loop closure updates.
These works have led to RGB-D scanning methods that feature robust, global tracking and can capture very large 3D environments.
However, although these methods can achieve stunning results in RGB-D capture and tracking, the quality of reconstructed 3D geometry nonetheless remains far from from artist-created 3D content, as the reconstructed scans are partial, and contain noise or oversmoothing from sensor quality or small camera tracking errors.

\paragraph{3D Features for Shape Alignment and Retrieval}

An alternative to bottom-up 3D reconstruction from RGB-D scanning techniques is to find high-quality CAD models that can replace the noisy and incomplete geometry from a 3D scan.
Finding and aligning these CAD models inevitably requires 3D feature descriptors to find geometric matches between the scan and the CAD models.
Traditionally, these descriptors were hand-crafted, and often based on a computation of histograms (e.g., point normals), such as FPFH~\cite{rusu2009fast}, SHOT~\cite{tombari2011combined}, or point-pair features~\cite{drost20123d}.
More recently, with advances in deep neural networks, these descriptors can be learned, for instance based on an implicit signed distance field representation \cite{zeng20173dmatch,deng2018ppf,deng2019direct}.
A typical pipeline for CAD-to-scan alignments builds on these descriptors; i.e., the first step is to find 3D feature matches and then use a variant of RANSAC or PnP to compute 6DoF or 9Dof CAD model alignments.
This two-step strategy has been used by Slam++~\cite{salas2013slam++}, Li et al.~\cite{li2015database}, Shao et al.~\cite{shao2012interactive}, but also by the data-driven work by Nan et al.~\cite{nan2012search} and the recent Scan2CAD approach~\cite{avetisyan2019scan2cad}.
Other approaches rely only on single RGB or RGB-D frame input, but use a similar two-step alignment strategy~\cite{lim2013parsing,izadinia2017im2cad,sun2018pix3d,huang2018holistic,gupta2015aligning,zou2018complete} images. 
While these methods are related, their focus is difference as we address geometric alignment independent of RGB information.

A fundamental limitation of these two-step pipelines is the decoupled nature of feature matching and alignment computation.
This inherently limits the ability of data-driven descriptors, as they remain unaware of the used optimization algorithm.
In our work, we propose an end-to-end alignment algorithm where correspondences are trained through gradients from an differentiable Procrustes optimizer.

\paragraph{Shape Retrieval Challenges and RGB-D Datasets}

In the context of 2D object alignment methods several datasets provide alignment annotations between RGB images and CAD models, including the PASCAL 3D+~\cite{xiang2014beyond}, 
ObjectNet3D~\cite{xiang2016objectnet3d},
the IKEA objects~\cite{lim2013parsing}, and Pix3D~\cite{sun2018pix3d}; however, no geometric information is given in the query images.

A very popular series of challenges in the context of shape retrieval is the SHREC, which is organized as part of Eurographics 3DOR~\cite{huashrec2017,phamshrec2018}; the tasks include matching object instances from ScanNet~\cite{dai2017scannet} and SceneNN~\cite{hua2016scenenn} to ShapeNet objects~\cite{shapenet2015}.

Scan2CAD~\cite{avetisyan2019scan2cad} is a very recent effort that provides accurate CAD alignment annotations on top of ScanNet~\cite{dai2017scannet} using ShapeNet models~\cite{shapenet2015}, based on roughly 100k manually annotated correspondences.
In addition to evaluating our method on the Scan2CAD test dataset, we also create an alignment benchmark on the synthetic SUNCG~\cite{song2017ssc} dataset.

\section{Overview}
The goal of our method is to align a set of CAD models to the objects of an input 3D scan.
That is, for an input 3D scan $\mathbb{S}$ of a scene, and a set of 3D CAD models $\mathbb{M}=\{m_i\}$, we aim to find 9DoF transformations $T_i$ (3 degrees each for translation, rotation, and scale) for each CAD model $m_i$ such that it aligns with a semantically matching object $\mathbb{O} = \{ o_j\}$ in $\mathbb{S}$.
This results in a complete, clean, CAD representation of the objects of a scene, as shown in Figure~\ref{fig:teaser}.

To this end, we propose an end-to-end 3D CNN-based approach to simultaneously retrieve and align CAD models to the objects of a scan in a single pass, for scans of varying sizes.
This end-to-end formulation enables the final alignment process to inform learning of scan-CAD correspondences.
To enable effective learning of scan-CAD object correspondences, we propose to use \emph{\CORRSFULL} (\CORRS{}), which establish dense correspondences between scan objects and CAD models, and are trained by our differentiable Procrustes alignment loss.

Then for an input scan $\mathbb{S}$ represented by volumetric grid encoding a truncated signed distance field, our model first detects object center locations as heatmap predictions over the volumetric grid and corresponding bounding box sizes for each object location. The bounding box represents the extent of the underlying object. 
From these detected object locations, we use the estimated bounding box size to crop out the neighborhood region around the object center from the learned feature space in order to predict our \CORR{} correspondences to CAD models.

From this neighborhood of feature information, we then predict \CORRS{}. 
These densely establish correspondences for each voxel in the object  neighborhood to CAD model space.
In order to be invariant to potential reflection and rotational symmetries, which could induce ambiguity in the correspondences, we simultaneously estimate the symmetry type of the object.
We additionally predict a binary mask to segment the object instance from background clutter in the neighborhood, thus informing the set of correspondences to be used for the final alignment.
To find a  CAD model corresponding to the scan object, we jointly learn an object descriptor which is used to retrieve a semantically similar CAD model from a database.

Finally, we introduce a differentiable Procrustes alignment, enabling an fully end-to-end formulation, where learned scan object-CAD model \CORR{} correspondences can be informed by the final alignment process, achieving efficient and accurate 9DoF CAD model alignment for 3D scans.

\section{Method}

\begin{figure*}[tp]
\begin{center}
   \includegraphics[width=\linewidth]{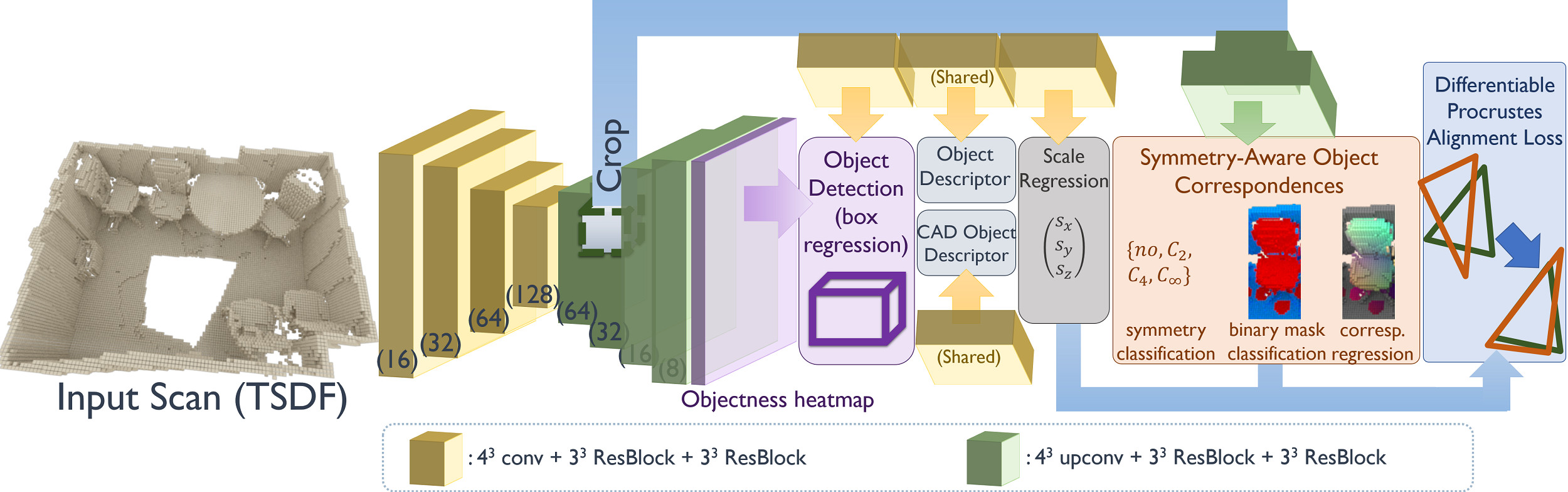}
    \end{center}
   \caption{Network architecture for our end-to-end approach for CAD model alignment. 
   An input TSDF scan represented in a volumetric grid is input to an encoder-decoder backbone constructed with residual blocks. 
   Objects are detected through objectness prediction and bounding box regression; these predicted object boxes are then used to crop features from the decoder to inform CAD model alignment to a detected object.
   The cropped features are processed to simultaneously predict an object descriptor constrained to be similar to a corresponding CAD object descriptor (used for retrieving CAD models), a $3$-dimensional scale, and our \CORRSFULL{} (\CORRS{}) which directly inform our differentiable Procrustes alignment loss.
   This enables correspondence learning to be informed by alignment, producing robust and efficient CAD model alignment.
   }
\label{fig:architecture}
\end{figure*}

\subsection{Network Architecture}
Our network architecture is shown in Figure~\ref{fig:architecture}.
It is designed to operate on 3D scans of varying sizes, in a fully-convolutional manner.
An input scan is given by a volumetric grid encoding a truncated signed distance field, representing the scan geometry.
To detect objects in a scan and align CAD models to them, we structure the network around a backbone, from which features can then extracted to predict individual \CORRS{}, informing the final alignment process.

The backbone of the network is structured in an encoder-decoder fashion, and composed of a series of ResNet blocks~\cite{he2016deep}. 
The bottleneck volume is spatially reduced by a factor of $16$ from the input volume, and is decoded to the original resolution through transpose convolutions. 
The decoder is structured symmetrically to the encoder, but with half the feature channels, which we empirically found to produce faster convergence and more accurate performance.
The output of the decoder is used to predict an objectness heatmap, identifying potential object locations, which is employed to inform bounding box regression for object detection.
The predicted object bounding boxes are used to crop and extract features from the output of the second decoder layer, which then inform the \CORRS{} predictions. 
The features used to inform the \CORR{} correspondence are extracted from the second block of the decoder, whose feature map spatial dimensions are $1/4$ of the original input dimension.  

\paragraph{Object Detection}

We first detect objects, predicting bounding boxes for the objects in a scan, which  then inform the \CORR{} predictions.
The output of the backbone decoder predicts heatmaps representing objectness probability over the full volumetric grid (whether the voxel is a center of an object). 
We then regress object bounding boxes corresponding to these potential object center locations.
For object bounding boxes predictions, we regress a $3$-channel feature map, with each $3$-dimensional vector corresponding to the bounding box extent size, and regressed using an $\ell_2$ loss.

Objectness is predicted as a heatmap, encoding voxel-wise probabilities as to whether each voxel is a center of an object.
To predict a location heatmap $H_1$, we additionally employ two proxy losses, using a second heatmap prediction $H_2$ as well as a predicted offset field $O$.
$H_1$ and $H_2$ are two $1$-channel heatmaps designed to encourage high recall and precision, respectively, and $O$ is a $3$-channel grid representing an offset field to the nearest object center.
The objectness heatmap loss is:
\begin{equation*}
    \mathcal{L}_{OD} = 2.0\cdot \mathcal{L}_{\textrm{recall}} + 10.0\cdot \mathcal{L}_{\textrm{precision}} + 10.0\cdot \mathcal{L}_{\textrm{offset}}
\end{equation*}
The weights for each component in the loss are designed to bring the losses numerically to approximately the same order of magnitude.

$\mathcal{L}_{\textrm{recall}}$ aims to achieve high recall. It operates on the prediction $H_1$, on which we apply a sigmoid activation and calculate the loss via binary-cross entropy (BCE). This loss on its own tends to establish a high recall, but also blurry predictions.
\begin{align}
\mathcal{L}_{\textrm{recall}} &= \sum_{x \in \Omega} \text{BCE}(\sigma(H_1(x)), H_\text{GT}(x)) \\
H_1 &: \Omega \rightarrow [0,1], \quad \sigma: \textrm{sigmoid} 
\end{align}

$\mathcal{L}_{\textrm{precision}}$ aims to achieve high precision. It operates on the prediction $H_2$, on which we apply a softmax activation and calculate the loss via negative log-likelihood (NLL).
Due to the softmax, this loss encourages highly localized predictions in the output volume, which helps to attain high precision.
\begin{align}
\mathcal{L}_{\textrm{recall}} &= \sum_{x \in \Omega} \text{NLL}(\sigma(H_2(x)), H_\text{GT}(x)) \\
H_2 &: \Omega \rightarrow [0,1], \quad \sigma: \textrm{softmax}
\end{align}

$\mathcal{L}_{\textrm{offset}}$ is a regression loss on the predicted a 3D offset field $O$, following \cite{papandreou2017towards}.
Each voxel of $O$ represents a $3$-dimensional vector that points to the nearest object center. This regression loss is used as a proxy loss to support the other two classification losses.
\begin{align}
\mathcal{L}_{\textrm{offset}} &= \sum_{x \in \Omega} \Vert O(x) -  O_\text{GT}(x) \Vert_2^2 \\
O &: \Omega \rightarrow \mathbb{R}^3 \notag
\end{align}

\paragraph{Predicting \CORRS{}}
\CORRS{} are dense, voxel-wise correspondences to CAD models. Hence, they are defined as $ \text{\CORR{}} : \Omega \rightarrow \Psi$ where  $\Psi$ depicts a closed space often as $\Psi \in [0,1]^3$ or in our case with ShapeNet $\Psi \in [-0.5,0.5]^3$; generally $\Psi$ depends on how the CAD models are normalized. In summary $\text{\CORR{}}(x_{\text{scan}})$ indicates the (normalized) coordinate in a CAD model of the correspondence to the given scan voxel $x_{\text{scan}}$.

\CORRS{} are predicted using features cropped from the network backbone.
For each detected object, we crop a region with the extend of the predicted bounding box volume $\mathcal{F}$ from the feature map of the second upsampling layer to inform our dense, symmetry-aware object correspondences. This feature volume $\mathcal{F}$ is first fitted through tri-linear interpolation into a uniform voxel grid of size  $48^3$ before streaming into different prediction heads. \CORRS{} incorporate several output predictions: a volume of dense correspondences from scan space to CAD object space, an instance segmentation mask, and a symmetry classification.

The dense correspondences, which map to CAD object space, implicitly contain CAD model alignment information. These correspondences are regressed as CAD object space coordinates, similar to \cite{wang2019normalized}, with the CAD object space defined as a uniform grid centered around the object, with coordinates normalized to $[-0.5,0.5]$. These coordinates are regressed using an $\ell_2$ loss.

In order to avoid ambiguities in correspondence that could be induced by object symmetries, we predict the symmetry class of the object for common symmetry classes for furniture objects: two-fold rotational symmetry, four-fold rotational symmetry, infinite rotational symmetry, and no symmetry.

Finally, to constrain the correspondences used for alignment to the scan object, we additionally predict a binary mask indicating the instance segmentation of the object, which is trained using a binary cross entropy loss.

\paragraph{Retrieval}

To retrieve a similar CAD model to the detected object, we use the cropped feature neighborhood $\mathcal{F}$ to train an object descriptor for the scan region, using a series of 3D convolutions to reduce the feature dimensionality to $8\times 4^3$.
This resulting $512$-dimensional object descriptor is then constrained to match the latent vector of an autoencoder trained on the CAD model dataset, with latent spaces constrained by an $\ell_2$ loss. This enables retrieval of a semantically similar CAD model at test time through a nearest neighbor search using the object descriptor.

\paragraph{9DoF Alignment}

Our differentiable 9DoF alignment enables training for CAD model alignment in an end-to-end fashion, thereby informing learned correspondences of the final alignment objective.
To this end, we leverage a differentiable Procrustes loss on the masked correspondences given by the \CORR{} predictions to find the rotation alignment. 
That is, we aim to find a rotation matrix $R$ which brings together the CAD and scan correspondence points $P_c, P_s$:
\begin{equation*}
    R = \textrm{argmin}_\Omega || \Omega P_c - P_s||_F,\quad\quad \Omega \in SO_3
\end{equation*}
This is solved through a differentiable SVD of $P_sP_c^T = U\Sigma V^T$, with $R=U \left[ \begin{smallmatrix} 1 & & \\ & 1 & \\ & & d\end{smallmatrix} \right]V^T$, $d = \text{det}(VU^T)$.
For scale and translation, we directly regress the scale using a series of 2 3D downsampling convolutions on $\mathcal{F}$, and the translation using the detected object center locations. 
Note that an object center is the geometric center of the bounding box. 

\subsection{Training}

\paragraph{Data}
Input scan data is represented by its truncated signed distance field (TSDF) encoded in a volumetric grid and generated through volumetric fusion~\cite{curless1996volumetric} (we use voxel size = $3$cm, truncation = $15$cm). 
The CAD models used to train the autoencoder to produce a latent space for scan object descriptor training are represented as unsigned distance fields (DF), using the level-set generation toolkit by Batty~\cite{battysdf}. 

To train our model for CAD model alignment for real scan data, we use the Scan2CAD dataset introduced by \cite{avetisyan2019scan2cad}. These Scan2CAD annotations provide 1506 scenes for training. 
Using upright rotation augmentation, we augment the number of training samples by 4 ($90^\circ$ increments with $20^\circ$ random jitter).
We train our network using full scenes as input, with batch size of 1. For \CORR{} prediction at train time the batch size is equal to the number of groundtruth objects in the given scene as crops are only performed around groundtruth object centers.
Only large scenes during training are randomly cropped to $400\times 400\times 64$ to meet memory requirements.
We found that training using 1 scene per batch generally yields stable convergence behavior.

For CAD model alignment to synthetic scan data, we use the SUNCG dataset~\cite{song2017ssc}, where we virtually scan the scenes following \cite{dai2018scancomplete,hou20193dsis} to produce input partial TSDF scans. 
The training process for synthetic SUNCG scan data is identical to training with real data. See supplemental material for further details.

\paragraph{Optimization}

We use an SGD optimizer with a batch size of 1 scene and an initial learning rate of 0.002, which is decayed by 0.5 every $20K$ iterations. We train for $50K$ iterations until convergence, which typically totals to 48 hours.

For object retrieval, we pre-train an autoencoder on all ShapeNetCore CAD models, trained to reconstruct their distance fields at $32^3$. 
This CAD autoencoder is trained with a batch size of 16 for $30K$ iterations.
We then train the full model with pre-calculated object descriptors for all ShapeNet models for CAD model alignment, with the CAD autoencoder latent space constraining the object descriptor training for retrieval.
\section{Results}
\begin{table*}[tp!]
\begin{center}
\footnotesize
\begin{tabular}{|l|r r r r r r r r r | r | r|}
\hline
   & bath             & bookshelf        & cabinet          & chair            & display          & sofa             & table            & trash bin        & other            & class avg.       & avg.           \\ \hline
\hline
FPFH (Rusu et al.~\cite{rusu2009fast}) & 0.00 & 1.92 & 0.00 & 10.00 & 0.00 & 5.41 & 2.04 & 1.75 & 2.00 & 2.57 & 4.45 \\ 
SHOT (Tombari et al.~\cite{tombari2010signature}) & 0.00 & 1.43 & 1.16 & 7.08 & 0.59 & 3.57 & 1.47 & 0.44 & 0.75 & 1.83 & 3.14 \\
Li et al.~\cite{li2015database} & 0.85 & 0.95 & 1.17 & 14.08 & 0.59 & 6.25 & 2.95 & 1.32 & 1.50 & 3.30 & 6.03 \\ 
3DMatch (Zeng et al.~\cite{zeng20173dmatch}) & 0.00 & 5.67 & 2.86 & 21.25 & 2.41 & 10.91 & 6.98 & 3.62 & 4.65 & 6.48 & 10.29  \\ 
Scan2CAD (Avetisyan et al.~\cite{avetisyan2019scan2cad})  & 36.20  & 36.40 & 34.00 & 44.26 & 17.89 & \textbf{70.63} & 30.66 & 30.11 & 20.60 & 35.64 & 31.68 \\ \hline
Direct 9DoF & 5.88 & 13.89 & 13.48 & 21.94 & 2.78 & 8.04 & 10.53 & 13.01 & 17.65 & 11.91 & 15.12 \\
Ours (no symmetry)        & 11.11 & 29.27 & 29.29 & 68.26 & 20.41 & 16.26 & 41.03 & 40.12 & 14.29 & 30 & 40.51 \\
Ours (no \CORRS{})        & 11.11 & 21.95 & 7.07 & 61.77 & 8.16 & 9.76 & 28.21 & 17.9 & 19.48 & 20.6 & 29.97 \\
Ours (no anchor)        & \textbf{45.24}          & \textbf{45.85} & 47.16     & 61.55   & \textbf{27.65}    & 51.92  & 41.21  & 31.13  & \textbf{29.62}     & 42.37 & 47.64 \\ 
Ours (no Procrustes)   & 33.33 & 36.59 & 28.28 & 50.51 & 14.29 & 13.01 & 58.97 & 35.19 & 28.57 & 33.19 & 35.74 \\
\textbf{Ours (final)}       & 38.89 & 41.46 & \textbf{51.52} & \textbf{73.04} & 26.53 & 26.83 & \textbf{76.92} & \textbf{48.15} & 18.18 & \textbf{44.61} & \textbf{50.72} \\
\hline
\end{tabular}
\end{center}
\caption{Accuracy comparison ($\%$) on Scan2CAD~\cite{avetisyan2019scan2cad}. 
We compare to state-of-the-art handcrafted feature descriptors (FPFH~\cite{rusu2009fast}, SHOT~\cite{tombari2010signature}, Li et al.~\cite{li2015database}) as well as learned descriptors (3DMatch~\cite{zeng20173dmatch}, Scan2CAD~\cite{avetisyan2019scan2cad}) for CAD model alignment.
These approaches consider correspondence finding and pose alignment optimization independently, while our end-to-end formulation can learn correspondences informed by alignment, achieving significantly higher CAD model alignment accuracy.
}
\label{tab:scannet_eval}
\end{table*}

\begin{figure*}[bp]
\begin{center}
\includegraphics[width=\linewidth]{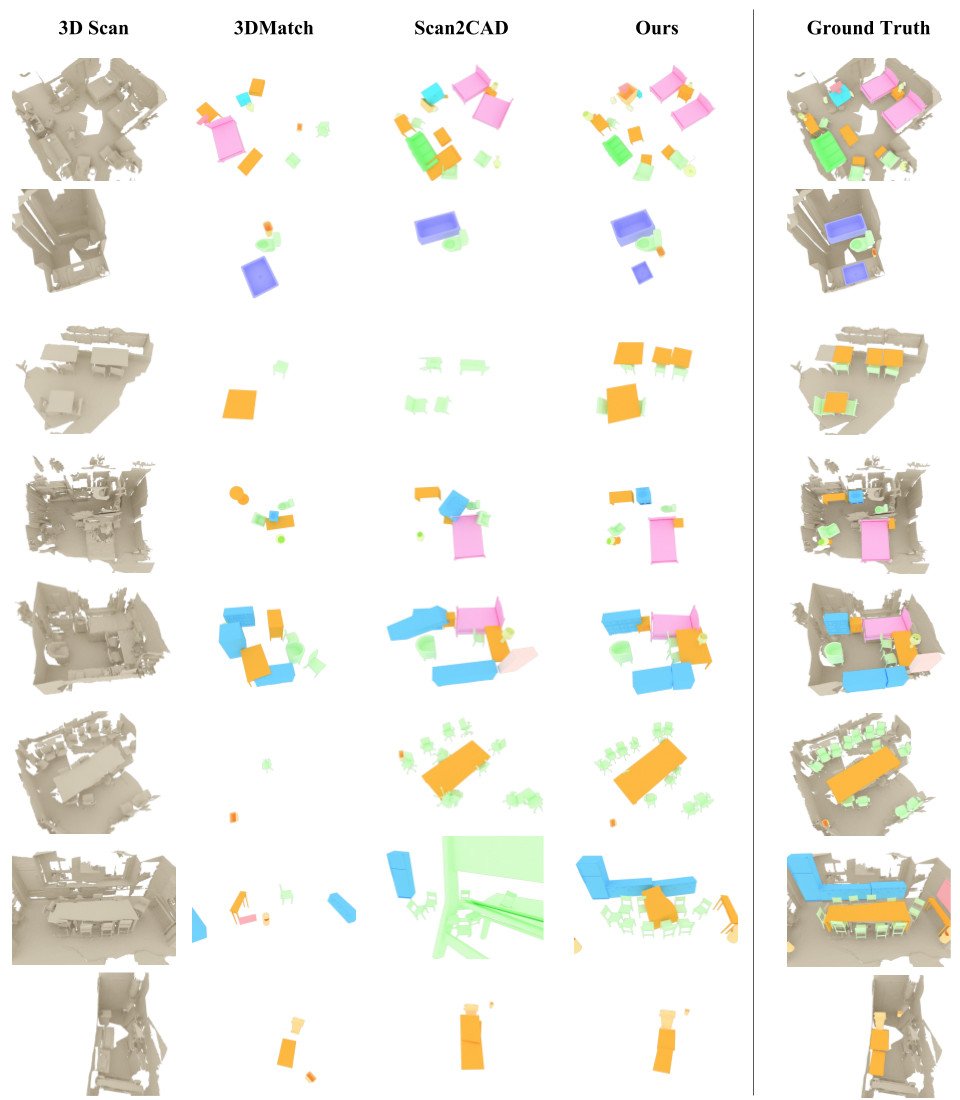}
\end{center}
\vspace{-0.2cm}
\caption{Qualitative comparison of CAD model alignment to ScanNet~\cite{dai2017scannet} scans. 
Our joint formulation of \CORR{} correspondence prediction and differentiable Procrustes alignment enable both more accurate and robust CAD model alignment estimation across varying scene types and sizes.
\vspace{-0.5cm}
}
\label{fig:results}
\end{figure*}

\begin{table}[tp]
\centering
     \resizebox{\columnwidth}{!}{
\begin{tabular}{|l|rrr|}
\hline
 Scene size        & small     & medium       & large \\
Scene dim         & $128 \times 96 \times 48$ & $144\times 128 \times 64$  & $256 \times 320\times 64$ \\
\# objects         & 7                 & 16                  & 20                 \\
\hline
Scan2CAD~\cite{avetisyan2019scan2cad} & 288.60s             & 565.86s              & 740.34s             \\
Ours & \textbf{0.62s}             & \textbf{1.11s}                & \textbf{2.60s}         \\
\hline
\end{tabular}
}
\caption{Runtime (in seconds) of our approach on varying-sized scenes for 9DoF CAD model alignment.
Our end-to-end approach predicts CAD model alignment in a single forward pass, enabling very efficient CAD model alignment -- several hundred times faster than previous data-driven approaches.}
\label{tab:timing}
\end{table}

We evaluate our proposed end-to-end approach for CAD model alignment in comparison to state of the art as well as with an ablation study analyzing our differentiable Procrustes alignment loss and various design choices.
We evaluate on real-world scans using the Scan2CAD dataset~\cite{avetisyan2019scan2cad}.
We use the evaluation metric proposed by Scan2CAD~\cite{avetisyan2019scan2cad}; that is, the ground truth CAD model pool is available as input, and a CAD model alignment is considered to be successful if the category of the CAD model matches that of the scan object and the alignment falls within $20$cm, $20^\circ$, and $20\%$ for translation, rotation, and scale, respectively.
For further evaluation on synthetic scans, we refer to the supplemental material.

\begin{figure*}[bp]
\begin{center}
\includegraphics[width=\linewidth]{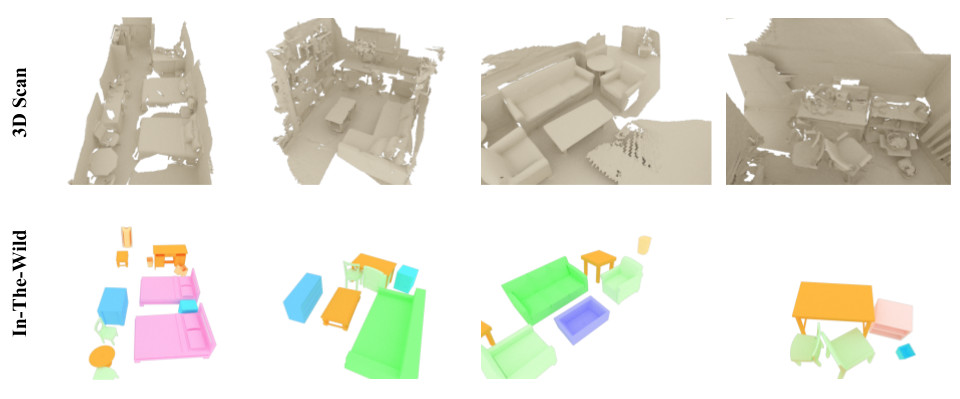}
\end{center}
\caption{
Our end-to-end CAD model alignment approach applied to an unconstrained set of candidate CAD models; here, we use a set of 3000 randomly selected CAD models from ShapeNetCore~\cite{shapenet2015}.
Our approach maintains robust CAD model alignment performance in such a  scenario which is often reflected in real-world applications.
}
\label{fig:results_unconstrained}
\end{figure*}

In addition to evaluating CAD model alignment using the Scan2CAD~\cite{avetisyan2019scan2cad} evaluation metrics, we also evaluate our approach on an unconstrained scenario with 3000 random CAD models as a candidate pool, shown in Figure~\ref{fig:results_unconstrained}.
In this scenario, we maintain robust CAD model alignment accuracy with a much larger set of possible CAD models.

\paragraph{Comparison to state of the art.}
Table~\ref{tab:scannet_eval} evaluates our approach against several state-of-the-art methods for CAD model alignment, which establish correspondences and alignment independently of each other. 
In particular, we compare to several approaches leveraging handcrafted feature descriptors: FPFH~\cite{rusu2009fast}, SHOT~\cite{tombari2011combined}, Li et al.~\cite{li2015database}, as well as learned feature descriptors: 3DMatch~\cite{zeng20173dmatch}, Scan2CAD~\cite{avetisyan2019scan2cad}. 
We follow these descriptors with RANSAC to obtain final alignment estimation, except for Scan2CAD, where we use the proposed alignment optimization.
Our end-to-end formulation, where correspondence learning can be informed by the alignment, outperforms these decoupled approaches by over $19.04\%$. 
Figure~\ref{fig:results} shows qualitative visualizations of our approach in comparison to these methods.

\paragraph{How much does the differentiable Procrustes alignment loss help?}
We additionally analyze the effect of our differentiable Procrustes loss.
In Table~\ref{tab:scannet_eval}, we compare several different alignment losses.
As a baseline, we train our model to directly regress the 9DoF alignment parameters with an $\ell_2$. We then evaluate our approach with (final) and without (no Procrustes) our differentiable Procrustes loss.
For CAD model alignment to 3D scans, our differentiable Procrustes alignment notably improves performance, by over $14.98\%$.

\paragraph{How much does \CORR{} prediction help?}
We evaluate our \CORR{} prediction on CAD model alignment in Table~\ref{tab:scannet_eval}. 
We train our model with (final) and without (no \CORRS{}) \CORR{} prediction as well as with coordinate correspondence prediction but without symmetry (no symmetry). We observe that our \CORR{} prediction significantly improves performance, by over $20.75\%$. 
Establishing \CORRS{} is fundamental to our approach, as dense correspondences can produce more reliable alignment, and unresolved symmetries can lead to ambiguities and inconsistencies in finding object correspondences.
In particular, we also evaluate the effect of symmetry classification in our \CORRS{}; explicitly predicting symmetry  yields a performance improvement of $10.21\%$.

\paragraph{What is the effect of using an anchor mechanism for object detection?}
In Table~\ref{tab:scannet_eval}, we also compare our CAD model alignment approach with (final) and without (no anchor) using anchors for object detection, where without anchors we predict only object center locations as a probability heatmap over the volumetric grid of the scan, but do not regress bounding boxes, and thus only crop a fixed neighborhood for the following \CORRS{} and alignment.
We observe that by employing  bounding box regression, we can  improve CAD model alignment performance, as this facilitates scale estimation and allows correspondence features to encompass the full object region.

\subsection{Limitations}

Although our approach shows significant improvements compared to state of the art, we believe there directions for improvement.
Currently, we focus on the objects in a scan, but do not consider structural components such as walls and floors.
We believe, however, that our method could be expanded to detect and match plane segments in the spirit of structural layout detection such as  PlaneRCNN~\cite{liu2018planercnn}.
In addition, we currently only consider the geometry of the scan or CAD; however, it is an interesting direction to consider finding matching textures in order to better visually match the appearance of a scan.
Finally, we hope to incorporate our alignment algorithm in an online system that can work at interactive rates and give immediate feedback to the scanning operator.
\section{Conclusion}
We have presented an end-to-end approach that automatically aligns CAD models with commodity 3D scans, which  that is facilitated with symmetry-aware correspondences and a differentiable Procrustes algorithm.
We show that by jointly training the correspondence prediction with direct, end-to-end alignment, our method is able to outperform existing state of the art by over $19.04\%$ in alignment accuracy.
In addition, our approach is roughly $250\times$ faster than previous data-driven approaches and thus could be easily incorporated into an online scanning system.
Overall, we believe that this is an important step towards obtaining clean and compact representations from 3D scans, and we hope it will open up future research in this direction.

\section*{Acknowledgements}
We would like to thank Justus Thies and J{\"u}rgen Sturm for valuable feedback.
This work is supported by Occipital, the ERC Starting Grant Scan2CAD (804724), a Google Faculty Award, an Nvidia Professorship Award, and the ZD.B.
We would also like to thank the support of the TUM-IAS, funded by the German Excellence Initiative and the European Union Seventh Framework Programme under grant agreement n° 291763, for the TUM-IAS Rudolf M{\"o}{\ss}bauer Fellowship
\clearpage
{\small
\bibliographystyle{ieee}
\bibliography{egbib}
}

\newpage
\begin{appendix}

\begin{table*}[tp!]
	\begin{center}
        \begin{center}
        \footnotesize
        \begin{tabular}{|l|r r r r r r r r r | r | r|}
        \hline
           & bath             & bookshelf        & cabinet          & chair            & display          & sofa             & table            & trash bin        & other            & class avg.       & avg.           \\ \hline
        \hline
        SHOT (Tombari et al.~\cite{tombari2010signature}) & 0 & 1.8 & 0 & 8.8 & 0.0 & 1.2 & 0 & 0 & 2.2 & 1.5 & 2.8 \\
        FPFH (Rusu et al.~\cite{rusu2009fast}) & 0 & 0 & 1.5 & 10.7 & 0 & 1.2 & 2.1 & 2.9 & 0 & 2.0 & 3.7 \\
        Li et al.~\cite{li2015database} & 0 & 1.8 & 2.3 & 1.11 & 0 & 2.8 & 6.4 & 2.7 & 0 & 3.0 & 4.6 \\ 
        3DMatch (Zeng et al.~\cite{zeng20173dmatch}) & 0 & 5.3 & 3.8 & 19.5 & 1.7 & 5.2 & 17.0 & 6.0 & 6.5 & 7.2 & 9.2  \\ 
        Scan2CAD (Avetisyan et al.~\cite{avetisyan2019scan2cad})  & 25.0 & 28.1 & 30.8 & 39.7 & 20.3 & 14.3 & 51.1 & 31.5 & 19.6 & 28.9 & 28.8 \\
        \textbf{Ours} & \textbf{40.6} & \textbf{38.6} & \textbf{36.2} & \textbf{68.1} & \textbf{25.4} & \textbf{27.0} & \textbf{63.8} & \textbf{38.0} & \textbf{40.2} & \textbf{42.0} & \textbf{44.1} \\
        \hline
        \end{tabular}
        \end{center}
        \captionof{table}{Performance comparison (\%) on the hidden test set of the Scan2CAD alignment benchmark~\cite{avetisyan2019scan2cad}. We outperform existing methods by a significant margin on all classes; the last two rows provide class and average instance alignment accuracy, respectively.}
        \label{tab:benchmark}
	\end{center}
	\end{table*}

\begin{figure*}[tb]
	\begin{center}
        \includegraphics[width=0.8\linewidth]{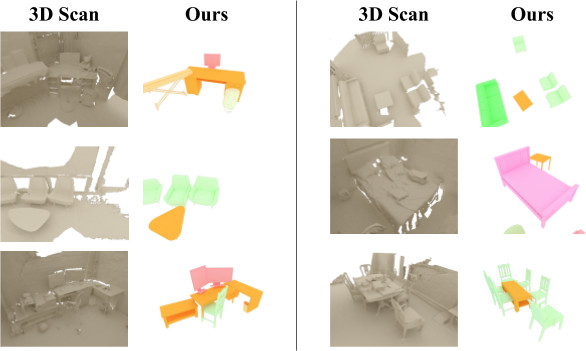}
        \captionof{figure}{Qualitative results on the Scan2CAD alignment benchmark~\cite{avetisyan2019scan2cad} (submitted to official benchmark website on March 29th, 2019)}
        \label{fig:results_benchmark}
	\end{center}
\end{figure*}

\section{Online Benchmark}
In this appendix, we provide additional results, including measurements on the hidden test set of the Scan2CAD benchmark~\cite{avetisyan2019scan2cad}.
Specifically, we provide a quantitative comparison in Tab.~\ref{tab:benchmark}, which was submitted to official benchmark website on March 29th, 2019.
In addition, we show qualitative results of our approach in Fig.~\ref{fig:results_benchmark}.

\section{SUNCG}
We conduct experiments on the SUNCG dataset~\cite{song2017ssc} to verify the effectiveness of our method. For training and evaluation, we create virtual scans of the synthetic scenes, where we simulate a large-scale indoor 3D reconstruction by using rendered depth frames similar to \cite{hou20193dsis,dai2018scancomplete} with the distinction that we add noise to the synthetic depth frames in the fusion process. The voxel resolution for the generated SDF grids is at $4.68cm$.
The ground truth models are provided by the SUNCG scenes, where we discard any objects that have not been seen during the virtual scanning (no occupancy in the scanned SDF). 
We show a quantitative evaluation in Tab.~\ref{tab:suncg_eval}, where we outperform the current state-of-the-art method Scan2CAD~\cite{avetisyan2019scan2cad} by a significant margin. We show that our method can align CAD models robustly through all classes. 
Additionally, we see that our Procrustes loss notably improves overall alignment accuracy. In particular, for less frequent CAD models (e.g., those summarized in \textit{other}), we observe a considerable improvement in alignment accuracy.

\begin{figure*}[tb!]
	\begin{center}
        \includegraphics[width=0.8\linewidth]{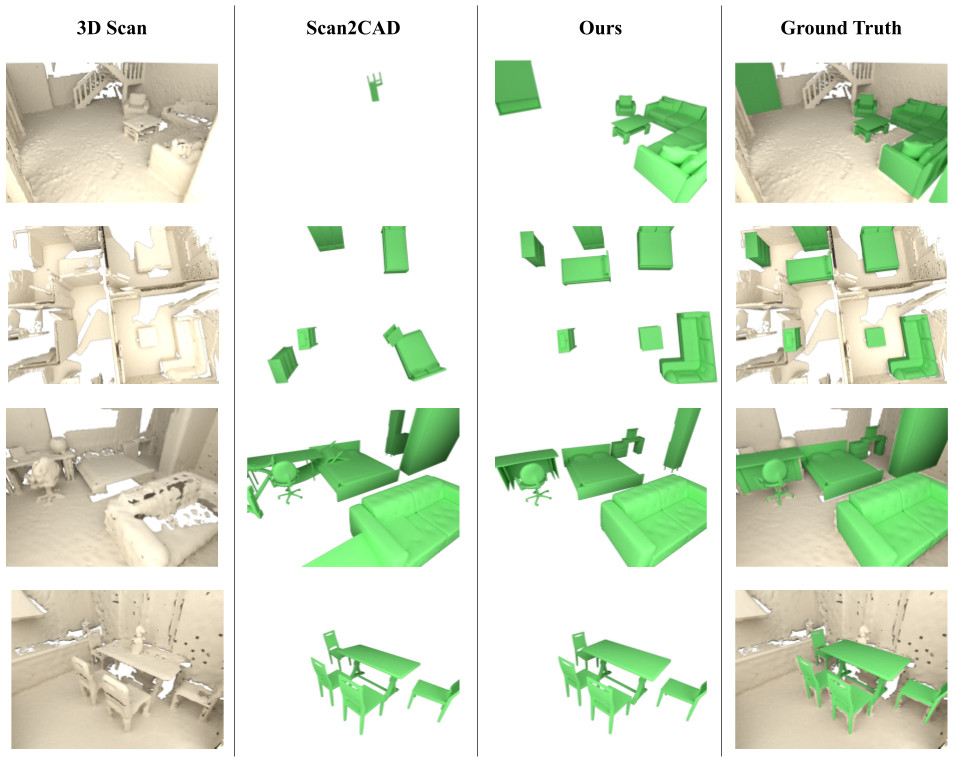}
        \captionof{figure}{Qualitative results on virtual scans from SUNCG. Note that our method handles complex CAD arrangements better than Scan2CAD.}
        \label{fig:results_suncg}
	\end{center}
\end{figure*}

Fig.~\ref{fig:results_suncg} shows qualitative results on scanned SUNCG scenes. Our end-to-end approach is able to handle large indoor scenes with complex furniture arrangements better than baseline methods.

\begin{table*}[tp!]
\begin{center}
\footnotesize
\begin{tabular}{|l|r r r r r r r r r | r | r|}
\hline
   & bed & cabinet & chair & desk & dresser & other & shelves & sofa & table  & class avg.       & avg.           \\ \hline
\hline
SHOT (Tombari et al.~\cite{tombari2010signature}) & 13.43 & 3.23 & 10.18 & 2.78 & 0 & 0 & 1.75 & 3.61 & 11.93 & 5.21 & 6.3\\
FPFH (Rusu et al.~\cite{rusu2009fast}) & 38.81 & 3.23 & 7.64 & 11.11 & 3.85 & 13.21 & 0 & 21.69 & 11.93 & 12.39 & 9.94  \\ 
Scan2CAD (Avetisyan et al.~\cite{avetisyan2019scan2cad})  & 52.24 & 17.97 & 36 & \textbf{30.56} & 3.85 & 20.75 & 7.89 & 40.96 & 43.12 & 28.15 & 29.23 \\ \hline
Ours (No Procrustes)       & \textbf{71.64} & 29.95 & 39.27 & 23.61 & 30.77 & 20.75 & 9.65 & \textbf{69.88} & 40.37 & 37.32 & 36.42 \\
\textbf{Ours (final)}       & \textbf{71.64} & \textbf{32.72} & \textbf{48.73} & 27.78 & \textbf{38.46} & \textbf{37.74} & \textbf{14.04} & 67.47 & \textbf{45.87} & \textbf{42.72} & \textbf{41.83} \\
\hline
\end{tabular}
\end{center}
\caption{CAD alignment accuracy comparison ($\%$) on SUNCG~\cite{song2017ssc}. 
We compare to state-of-the-art handcrafted feature descriptors FPFH~\cite{rusu2009fast}, SHOT~\cite{tombari2010signature} as well as a learning based method Scan2CAD~\cite{avetisyan2019scan2cad} for CAD model alignment. Note that the Procrustes loss considerably improves overall alignment accuracy.
}
\label{tab:suncg_eval}
\end{table*}

\end{appendix}

\end{document}